\documentclass[10pt,twocolumn,letterpaper]{article}

\usepackage{cvpr}
\usepackage{times}
\usepackage{epsfig}
\usepackage{graphicx}
\usepackage{amsmath}
\usepackage{amssymb}

\usepackage{url}
\usepackage{multirow}

\newcommand{\specialcell}[2][c]{\begin{tabular}[#1]{@{}c@{}}#2\end{tabular}}

\usepackage[pagebackref=true,breaklinks=true,letterpaper=true,colorlinks,bookmarks=false]{hyperref}

\cvprfinalcopy 

\def\cvprPaperID{1867} 

\ifcvprfinal\fi
\begin{document}

\title{Quantized Convolutional Neural Networks for Mobile Devices}
\author{
  Jiaxiang Wu, Cong Leng, Yuhang Wang, Qinghao Hu, Jian Cheng\\
  National Laboratory of Patter Recognition\\
  Institute of Automation, Chinese Academy of Sciences\\
  {\tt\small \{jiaxiang.wu, cong.leng, yuhang.wang, qinghao.hu, jcheng\}@nlpr.ia.ac.cn}
}

\maketitle


\begin{abstract}
Recently, convolutional neural networks (CNN) have demonstrated impressive performance in various computer vision tasks. However, high performance hardware is typically indispensable for the application of CNN models due to the high computation complexity, which prohibits their further extensions. In this paper, we propose an efficient framework, namely Quantized CNN, to simultaneously speed-up the computation and reduce the storage and memory overhead of CNN models. Both filter kernels in convolutional layers and weighting matrices in fully-connected layers are quantized, aiming at minimizing the estimation error of each layer's response. Extensive experiments on the ILSVRC-12 benchmark demonstrate $4 \sim 6 \times$ speed-up and $15 \sim 20 \times$ compression with merely one percentage loss of classification accuracy. With our quantized CNN model, even mobile devices can accurately classify images within one second.
\end{abstract}


\section{Introduction}

\begin{figure}[!ht]
\centering
\includegraphics[width = \linewidth]{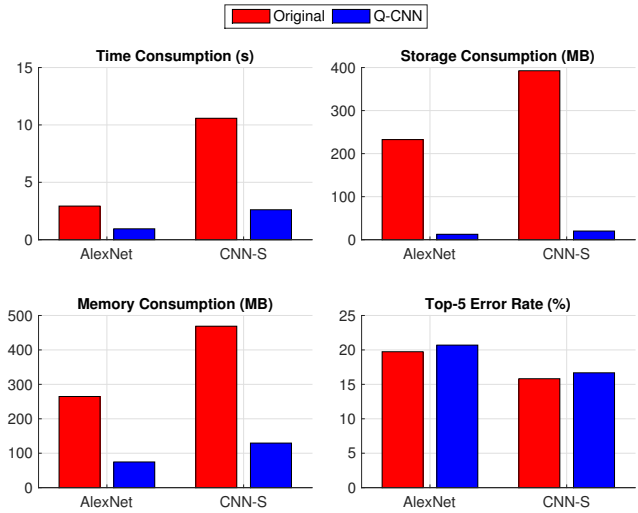}
\caption{Comparison on the efficiency and classification accuracy between the original and quantized AlexNet \cite{krizhevsky2012imagenet} and CNN-S \cite{chatfield2014return} on a Huawei\textsuperscript{\textregistered} Mate 7 smartphone.}
\label{fig:mobile_devices}
\end{figure}

In recent years, we have witnessed the great success of convolutional neural networks (CNN) \cite{lecun1989backpropagation} in a wide range of visual applications, including image classification \cite{krizhevsky2012imagenet, simonyan2015very}, object detection \cite{girshick2014rich, girshick2015fast}, age estimation \cite{li2012learning, levi2015age}, etc. This success mainly comes from deeper network architectures as well as the tremendous training data. However, as the network grows deeper, the model complexity is also increasing exponentially in both the training and testing stages, which leads to the very high demand in the computation ability. For instance, the 8-layer AlexNet \cite{krizhevsky2012imagenet} involves 60M parameters and requires over 729M FLOPs\footnotemark to classify a single image. Although the training stage can be offline carried out on high performance clusters with GPU acceleration, the testing computation cost may be unaffordable for common personal computers and mobile devices. Due to the limited computation ability and memory space, mobile devices are almost intractable to run deep convolutional networks. Therefore, it is crucial to accelerate the computation and compress the memory consumption for CNN models.

For most CNNs, convolutional layers are the most time-consuming part, while fully-connected layers involve massive network parameters. Due to the intrinsical difference between them, existing works usually focus on improving the efficiency for either convolutional layers or fully-connected layers. In \cite{denton2014exploiting, jaderberg2014speeding, zhang2015efficient, zhang2015accelerating, lebedev2015fast, lebedev2015speeding}, low-rank approximation or tensor decomposition is adopted to speed-up convolutional layers. On the other hand, parameter compression in fully-connected layers is explored in \cite{ciresan2011high, denton2014exploiting, gong2014compressing, yang2014deep, chen2015compressing_icml, han2015deep, srinivas2015data}. Overall, the above-mentioned algorithms are able to achieve faster speed or less storage. However, few of them can achieve significant acceleration and compression simultaneously for the whole network.
\footnotetext{FLOPs: number of FLoating-point OPerations required to classify one image with the convolutional network.}

In this paper, we propose a unified framework for convolutional networks, namely Quantized CNN (Q-CNN), to simultaneously accelerate and compress CNN models with only minor performance degradation. With network parameters quantized, the response of both convolutional and fully-connected layers can be efficiently estimated via the approximate inner product computation. We minimize the estimation error of each layer's response during parameter quantization, which can better preserve the model performance. In order to suppress the accumulative error while quantizing multiple layers, an effective training scheme is introduced to take previous estimation error into consideration. Our Q-CNN model enables fast test-phase computation, and the storage and memory consumption are also significantly reduced.

We evaluate our Q-CNN framework for image classification on two benchmarks, MNIST \cite{lecun1998gradient} and ILSVRC-12 \cite{russakovsky2015imagenet}. For MNIST, our Q-CNN approach achieves over 12$\times$ compression for two neural networks (no convolution), with lower accuracy loss than several baseline methods. For ILSVRC-12, we attempt to improve the test-phase efficiency of four convolutional networks: AlexNet \cite{krizhevsky2012imagenet}, CaffeNet \cite{jia2014caffe}, CNN-S \cite{chatfield2014return}, and VGG-16 \cite{simonyan2015very}. Generally, Q-CNN achieves 4$\times$ acceleration and $15\times$ compression (sometimes higher) for each network, with less than 1\% drop in the top-5 classification accuracy. Moreover, we implement the quantized CNN model on mobile devices, and dramatically improve the test-phase efficiency, as depicted in Figure \ref{fig:mobile_devices}. The main contributions of this paper can be summarized as follows:
\begin{itemize}
\setlength{\itemsep}{0pt}
\setlength{\parskip}{0pt}
\setlength{\parsep}{0pt}
\item We propose a unified Q-CNN framework to accelerate and compress convolutional networks. We demonstrate that better quantization can be learned by minimizing the estimation error of each layer's response.
\item We propose an effective training scheme to suppress the accumulative error while quantizing the whole convolutional network.
\item Our Q-CNN framework achieves $4 \sim 6 \times$ speed-up and $15 \sim 20 \times$ compression, while the classification accuracy loss is within one percentage. Moreover, the quantized CNN model can be implemented on mobile devices and classify an image within one second.
\end{itemize}

\section{Preliminary}

During the test phase of convolutional networks, the computation overhead is dominated by convolutional layers; meanwhile, the majority of network parameters are stored in fully-connected layers. Therefore, for better test-phase efficiency, it is critical to speed-up the convolution computation and compress parameters in fully-connected layers.

Our observation is that the forward-passing process of both convolutional and fully-connected layers is dominated by the computation of inner products. More formally, we consider a convolutional layer with input feature maps $S \in \mathbb{R}^{d_{s} \times d_{s} \times C_{s}}$ and response feature maps $T \in \mathbb{R}^{d_{t} \times d_{t} \times C_{t}}$, where $d_{s}, d_{t}$ are the spatial sizes and $C_{s}, C_{t}$ are the number of feature map channels. The response at the 2-D spatial position $p_{t}$ in the $c_{t}$-th response feature map is computed as:
\begin{equation}
T_{p_{t}} ( c_{t} ) = \sum\nolimits_{( p_{k}, p_{s} )} \langle W_{c_{t}, p_{k}}, S_{p_{s}} \rangle
\end{equation}
where $W_{c_{t}} \in \mathbb{R}^{d_{k} \times d_{k} \times C_{s}}$ is the $c_{t}$-th convolutional kernel and $d_{k}$ is the kernel size. We use $p_{s}$ and $p_{k}$ to denote the 2-D spatial positions in the input feature maps and convolutional kernels, and both $W_{c_{t}, p_{k}}$ and $S_{p_{s}}$ are $C_{s}$-dimensional vectors. The layer response is the sum of inner products at all positions within the $d_{k} \times d_{k}$ receptive field in the input feature maps.

Similarly, for a fully-connected layer, we have:
\begin{equation}
T ( c_{t} ) = \langle W_{c_{t}}, S \rangle
\end{equation}
where $S \in \mathbb{R}^{C_{s}}$ and $T \in \mathbb{R}^{C_{t}}$ are the layer input and layer response, respectively, and $W_{c_{t}} \in \mathbb{R}^{C_{s}}$ is the weighting vector for the $c_{t}$-th neuron of this layer.

Product quantization \cite{jegou2011product} is widely used in approximate nearest neighbor search, demonstrating better performance than hashing-based methods \cite{leng2015online, leng2015hashing}. The idea is to decompose the feature space as the Cartesian product of multiple subspaces, and then learn sub-codebooks for each subspace. A vector is represented by the concatenation of sub-codewords for efficient distance computation and storage.

In this paper, we leverage product quantization to implement the efficient inner product computation. Let us consider the inner product computation between $x, y \in \mathbb{R}^{D}$. At first, both $x$ and $y$ are split into $M$ sub-vectors, denoted as $x^{( m )}$ and $y^{( m )}$. Afterwards, each $x^{( m )}$ is quantized with a sub-codeword from the $m$-th sub-codebook, then we have
\begin{equation}
\langle y, x \rangle = \sum\nolimits_{m} \langle y^{( m )}, x^{( m )} \rangle \approx \sum\nolimits_{m} \langle y^{( m )}, c_{k_{m}}^{( m )} \rangle
\end{equation}
which transforms the $\mathcal{O} ( D )$ inner product computation to $M$ addition operations ($M \leq D$), if the inner products between each sub-vector $y^{( m )}$ and all the sub-codewords in the $m$-th sub-codebook have been computed in advance.

Quantization-based approaches have been explored in several works \cite{gong2014compressing, chen2015compressing_icml, han2015deep}. These approaches mostly focus on compressing parameters in fully-connected layers \cite{gong2014compressing, chen2015compressing_icml}, and none of them can provide acceleration for the test-phase computation. Furthermore, \cite{gong2014compressing, han2015deep} require the network parameters to be re-constructed during the test-phase, which limit the compression to disk storage instead of memory consumption. On the contrary, our approach offers simultaneous acceleration and compression for both convolutional and fully-connected layers, and can reduce the run-time memory consumption dramatically.

\section{Quantized CNN}

In this section, we present our approach for accelerating and compressing convolutional networks. Firstly, we introduce an efficient test-phase computation process with the network parameters quantized. Secondly, we demonstrate that better quantization can be learned by directly minimizing the estimation error of each layer's response. Finally, we analyze the computation complexity of our quantized CNN model.

\subsection{Quantizing the Fully-connected Layer}

For a fully-connected layer, we denote its weighting matrix as $W \in \mathbb{R}^{C_{s} \times C_{t}}$, where $C_{s}$ and $C_{t}$ are the dimensions of the layer input and response, respectively. The weighting vector $W_{c_{t}}$ is the $c_{t}$-th column vector in $W$.

We evenly split the $C_{s}$-dimensional space (where $W_{c_{t}}$ lies in) into $M$ subspaces, each of $C'_{s} = C_{s} / M$ dimensions. Each $W_{c_{t}}$ is then decomposed into $M$ sub-vectors, denoted as $W_{c_{t}}^{( m )}$. A sub-codebook can be learned for each subspace after gathering all the sub-vectors within this subspace. Formally, for the $m$-th subspace, we optimize:
\begin{equation}
\begin{split}
\min\limits_{D^{( m )}, B^{( m )}} &~ \left\| D^{( m )} B^{( m )} - W^{( m )} \right\|_{F}^{2} \\
s.t. &~ D^{( m )} \in \mathbb{R}^{C'_{s} \times K}, B^{( m )} \in \{ 0, 1 \}^{K \times C_{t}}
\end{split}
\end{equation}
where $W^{( m )} \in \mathbb{R}^{C'_{s} \times C_{t}}$ consists of the $m$-th sub-vectors of all weighting vectors. The sub-codebook $D^{( m )}$ contains $K$ sub-codewords, and each column in $B^{( m )}$ is an indicator vector (only one non-zero entry), specifying which sub-codeword is used to quantize the corresponding sub-vector. The optimization can be solved via k-means clustering.

The layer response is approximately computed as:
\small
\begin{equation}
\begin{split}
T ( c_{t} ) = \sum\nolimits_{m} \langle W_{c_{t}}^{( m )}, S^{( m )} \rangle &\approx \sum\nolimits_{m} \langle D^{( m )} B_{c_{t}}^{( m )}, S^{( m )} \rangle \\
&= \sum\nolimits_{m} \langle D_{k_{m} ( c_{t} )}^{( m )}, S^{( m )} \rangle
\end{split}
\end{equation}
\normalsize
where $B_{c_{t}}^{( m )}$ is the $c_{t}$-th column vector in $B^{( m )}$, and $S^{( m )}$ is the $m$-th sub-vector of the layer input. $k_{m} ( c_{t} )$ is the index of the sub-codeword used to quantize the sub-vector $W_{c_{t}}^{( m )}$.

\begin{figure}[!ht]
\centering
\includegraphics[width = \linewidth]{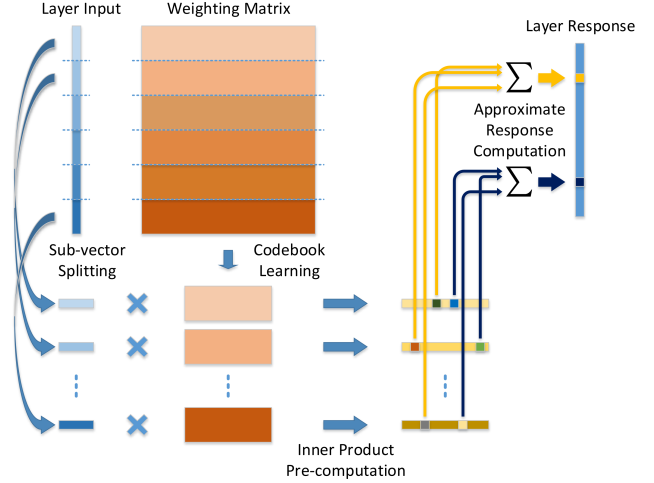}
\caption{The parameter quantization and test-phase computation process of the fully-connected layer.}
\label{fig:pq_for_fcnt}
\end{figure}

In Figure \ref{fig:pq_for_fcnt}, we depict the parameter quantization and test-phase computation process of the fully-connected layer. By decomposing the weighting matrix into $M$ sub-matrices, $M$ sub-codebooks can be learned, one per subspace. During the test-phase, the layer input is split into $M$ sub-vectors, denoted as $S^{( m )}$. For each subspace, we compute the inner products between $S^{( m )}$ and every sub-codeword in $D^{( m )}$, and store the results in a look-up table. Afterwards, only $M$ addition operations are required to compute each response. As a result, the overall time complexity can be reduced from $\mathcal{O} ( C_{s} C_{t} )$ to $\mathcal{O} ( C_{s} K + C_{t} M )$. On the other hand, only sub-codebooks and quantization indices need to be stored, which can dramatically reduce the storage consumption.

\subsection{Quantizing the Convolutional Layer}

Unlike the 1-D weighting vector in the fully-connected layer, each convolutional kernel is a 3-dimensional tensor: $W_{c_{t}} \in \mathbb{R}^{d_{k} \times d_{k} \times C_{s}}$. Before quantization, we need to determine how to split it into sub-vectors, i.e. apply subspace splitting to which dimension. During the test phase, the input feature maps are traversed by each convolutional kernel with a sliding window in the spatial domain. Since these sliding windows are partially overlapped, we split each convolutional kernel along the dimension of feature map channels, so that the pre-computed inner products can be re-used at multiple spatial locations. Specifically, we learn the quantization in each subspace by:
\begin{equation}
\begin{split}
\min\limits_{D^{( m )}, \{ B_{p_{k}}^{( m )} \}} &~ \sum\nolimits_{p_{k}} \left\| D^{( m )} B_{p_{k}}^{( m )} - W_{p_{k}}^{( m )} \right\|_{F}^{2} \\
s.t. &~ D^{( m )} \in \mathbb{R}^{C'_{s} \times K}, B_{p_{k}}^{( m )} \in \{ 0, 1 \}^{K \times C_{t}}
\end{split}
\end{equation}
where $W_{p_{k}}^{( m )} \in \mathbb{R}^{C'_{s} \times C_{t}}$ contains the $m$-th sub-vectors of all convolutional kernels at position $p_{k}$. The optimization can also be solved by k-means clustering in each subspace.

With the convolutional kernels quantized, we approximately compute the response feature maps by:
\begin{equation}
\begin{split}
T_{p_{t}} ( c_{t} ) &= \sum\nolimits_{( p_{k}, p_{s} )} \sum\nolimits_{m} \langle W_{c_{t}, p_{k}}^{( m )}, S_{p_{s}}^{( m )} \rangle \\
&\approx \sum\nolimits_{( p_{k}, p_{s} )} \sum\nolimits_{m} \langle D^{( m )} B_{c_{t}, p_{k}}^{( m )}, S_{p_{s}}^{( m )} \rangle \\
&= \sum\nolimits_{( p_{k}, p_{s} )} \sum\nolimits_{m} \langle D_{k_{m} ( c_{t}, p_{k} )}^{( m )}, S_{p_{s}}^{( m )} \rangle
\end{split}
\label{eqn:comp_conv_response}
\end{equation}
where $S_{p_{s}}^{( m )}$ is the $m$-th sub-vector at position $p_{s}$ in the input feature maps, and $k_{m} ( c_{t}, p_{k} )$ is the index of the sub-codeword to quantize the $m$-th sub-vector at position $p_{k}$ in the $c_{t}$-th convolutional kernel.

Similar to the fully-connected layer, we pre-compute the look-up tables of inner products with the input feature maps. Then, the response feature maps are approximately computed with (\ref{eqn:comp_conv_response}), and both the time and storage complexity can be greatly reduced.

\subsection{Quantization with Error Correction}

So far, we have presented an intuitive approach to quantize parameters and improve the test-phase efficiency of convolutional networks. However, there are still two critical drawbacks. First, minimizing the quantization error of model parameters does not necessarily give the optimal quantized network for the classification accuracy. In contrast, minimizing the estimation error of each layer's response is more closely related to the network's classification performance. Second, the quantization of one layer is independent of others, which may lead to the accumulation of error when quantizing multiple layers. The estimation error of the network's final response is very likely to be quickly accumulated, since the error introduced by the previous quantized layers will also affect the following layers.

To overcome these two limitations, we introduce the idea of error correction into the quantization of network parameters. This improved quantization approach directly minimizes the estimation error of the response at each layer, and can compensate the error introduced by previous layers. With the error correction scheme, we can quantize the network with much less performance degradation than the original quantization method.

\subsubsection{Error Correction for the Fully-connected Layer}

Suppose we have $N$ images to learn the quantization of a fully-connected layer, and the layer input and response of image $I_{n}$ are denoted as $S_{n}$ and $T_{n}$. In order to minimize the estimation error of the layer response, we optimize:
\begin{equation}
\min\limits_{\{ D^{( m )} \}, \{ B^{( m )} \}} \sum\nolimits_{n} \left\| T_{n} - \sum\nolimits_{m} ( D^{( m )} B^{( m )} )^{T} S_{n}^{( m )} \right\|_{F}^{2}
\label{eqn:ecpq_fcnt_obj_func}
\end{equation}
where the first term in the Frobenius norm is the desired layer response, and the second term is the approximated layer response computed via the quantized parameters.

A block coordinate descent approach can be applied to minimize this objective function. For the $m$-th subspace, its residual error is defined as:
\begin{equation}
R_{n}^{( m )} = T_{n} - \sum\nolimits_{m' \neq m} ( D^{( m' )} B^{( m' )} )^{T} S_{n}^{( m' )}
\end{equation}
and then we attempt to minimize the residual error of this subspace, which is:
\begin{equation}
\min\limits_{D^{( m )}, B^{( m )}} \sum\nolimits_{n} \left\| R_{n}^{( m )} - ( D^{( m )} B^{( m )} )^{T} S_{n}^{( m )} \right\|_{F}^{2}
\label{eqn:optm_ecpq_fcnt}
\end{equation}
and the above optimization can be solved by alternatively updating the sub-codebook and sub-codeword assignment.

\textbf{Update \boldmath$D^{( m )}$.} We fix the sub-codeword assignment $B^{( m )}$, and define $L_{k} = \{ c_{t} | B^{( m )} ( k, c_{t} ) = 1 \}$. The optimization in (\ref{eqn:optm_ecpq_fcnt}) can be re-formulated as:
\begin{equation}
\min\limits_{\{ D_{k}^{( m )} \}} \sum\nolimits_{n, k} \sum\nolimits_{c_{t} \in L_{k} } [ R_{n}^{( m )} ( c_{t} ) - D_{k}^{( m )^{T}} S_{n}^{( m )} ]^{2}
\label{eqn:optm_ecpq_fcnt_dcomp}
\end{equation}
which implies that the optimization over one sub-codeword does not affect other sub-codewords. Hence, for each sub-codeword, we construct a least square problem from (\ref{eqn:optm_ecpq_fcnt_dcomp}) to update it.

\textbf{Update \boldmath$B^{( m )}$.} With the sub-codebook $D^{( m )}$ fixed, it is easy to discover that the optimization of each column in $B^{( m )}$ is mutually independent. For the $c_{t}$-th column, its optimal sub-codeword assignment is given by:
\begin{equation}
k_{m}^{*} ( c_{t} ) = \arg\min_{k} \sum\nolimits_{n} [ R_{n}^{( m )} ( c_{t} ) - D_{k}^{( m )^{T}} S_{n}^{( m )} ]^{2}
\end{equation}

\subsubsection{Error Correction for the Convolutional Layer}
\label{sss:err_corr_conv}

We adopt the similar idea to minimize the estimation error of the convolutional layer's response feature maps, that is:
\small
\begin{equation}
\min\limits_{\{ D^{( m )} \}, \{ B_{p_{k}}^{( m )} \}} \sum_{n, p_{t}} \left\| T_{n, p_{t}} - \sum_{( p_{k}, p_{s} )} \sum_{m} ( D^{( m )} B_{p_{k}}^{( m )} )^{T} S_{n, p_{s}}^{( m )} \right\|_{F}^{2}
\label{eqn:ecpq_conv_obj_func}
\end{equation}
\normalsize

The optimization also can be solved by block coordinate descent. More details on solving this optimization can be found in the supplementary material.

\subsubsection{Error Correction for Multiple Layers}

The above quantization method can be sequentially applied to each layer in the CNN model. One concern is that the estimation error of layer response caused by the previous layers will be accumulated and affect the quantization of the following layers. Here, we propose an effective training scheme to address this issue.

We consider the quantization of a specific layer, assuming its previous layers have already been quantized. The optimization of parameter quantization is based on the layer input and response of a group of training images. To quantize this layer, we take the layer input in the quantized network as $\{ S_{n} \}$, and the layer response in the original network (not quantized) as $\{ T_{n} \}$ in Eq. (\ref{eqn:ecpq_fcnt_obj_func}) and (\ref{eqn:ecpq_conv_obj_func}). In this way, the optimization is guided by the actual input in the quantized network and the desired response in the original network. The accumulative error introduced by the previous layers is explicitly taken into consideration during optimization. In consequence, this training scheme can effectively suppress the accumulative error for the quantization of multiple layers.

Another possible solution is to adopt back-propagation to jointly update the sub-codebooks and sub-codeword assignments in all quantized layers. However, since the sub-codeword assignments are discrete, the gradient-based optimization can be quite difficult, if not entirely impossible. Therefore, back-propagation is not adopted here, but could be a promising extension for future work.

\subsection{Computation Complexity}

Now we analyze the test-phase computation complexity of convolutional and fully-connected layers, with or without parameter quantization. For our proposed Q-CNN model, the forward-passing through each layer mainly consists of two procedures: pre-computation of inner products, and approximate computation of layer response. Both sub-codebooks and sub-codeword assignments are stored for the test-phase computation. We report the detailed comparison on the computation and storage overhead in Table \ref{tab:comp_comp_stor_overhead}.

\begin{table}[!ht]
\centering
\caption{Comparison on the computation and storage overhead of convolutional and fully-connected layers.}
\begin{tabular}{c|c|c|c}
\hline
\multirow{4}{*}{FLOPs} & \multirow{2}{*}{Conv.} & CNN & $d_{t}^{2} C_{t} d_{k}^{2} C_{s}$ \\ \cline{3-4}
 & & Q-CNN & $d_{s}^{2} C_{s} K +  d_{t}^{2} C_{t} d_{k}^{2} M$ \\ \cline{2-4}
 & \multirow{2}{*}{FCnt.} & CNN & $C_{s} C_{t}$ \\ \cline{3-4}
 & & Q-CNN & $C_{s} K + C_{t} M$ \\ \hline \hline
\multirow{4}{*}{Bytes} & \multirow{2}{*}{Conv.} & CNN & $4 d_{k}^{2} C_{s} C_{t}$ \\ \cline{3-4}
 & & Q-CNN & $4 C_{s} K + \tfrac{1}{8} d_{k}^{2} M C_{t} \log_{2} K$ \\ \cline{2-4}
 & \multirow{2}{*}{FCnt.} & CNN & $4 C_{s} C_{t}$ \\ \cline{3-4}
 & & Q-CNN & $4 C_{s} K + \tfrac{1}{8} M C_{t} \log_{2} K$ \\ \hline
\end{tabular}
\label{tab:comp_comp_stor_overhead}
\end{table}

As we can see from Table \ref{tab:comp_comp_stor_overhead}, the reduction in the computation and storage overhead largely depends on two hyper-parameters, $M$ (number of subspaces) and $K$ (number of sub-codewords in each subspace). Large values of $M$ and $K$ lead to more fine-grained quantization, but is less efficient in the computation and storage consumption. In practice, we can vary these two parameters to balance the trade-off between the test-phase efficiency and accuracy loss of the quantized CNN model.

\section{Related Work}

There have been a few attempts in accelerating the test-phase computation of convolutional networks, and many are inspired from the low-rank decomposition. Denton et al. \cite{denton2014exploiting} presented a series of low-rank decomposition designs for convolutional kernels. Similarly, CP-decomposition was adopted in \cite{lebedev2015speeding} to transform a convolutional layer into multiple layers with lower complexity. Zhang et al. \cite{zhang2015efficient, zhang2015accelerating} considered the subsequent nonlinear units while learning the low-rank decomposition. \cite{lebedev2015fast} applied group-wise pruning to the convolutional tensor to decompose it into the multiplications of thinned dense matrices. Recently, fixed-point based approaches are explored in \cite{courbariaux2015training, rastegari2016xnor}. By representing the connection weights (or even network activations) with fixed-point numbers, the computation can greatly benefit from hardware acceleration.

Another parallel research trend is to compress parameters in fully-connected layers. Ciresan et al. \cite{ciresan2011high} randomly remove connection to reduce network parameters. Matrix factorization was adopted in \cite{denil2013predicting, denton2014exploiting} to decompose the weighting matrix into two low-rank matrices, which demonstrated that significant redundancy did exist in network parameters. Hinton et al. \cite{hinton2015distilling} proposed to use dark knowledge (the response of a well-trained network) to guide the training of a much smaller network, which was superior than directly training. By exploring the similarity among neurons, Srinivas et al. \cite{srinivas2015data} proposed a systematic way to remove redundant neurons instead of network connections. In \cite{yang2014deep}, multiple fully-connected layers were replaced by a single ``Fastfood'' layer, which can be trained in an end-to-end style with convolutional layers. Chen et al. \cite{chen2015compressing_icml} randomly grouped connection weights into hash buckets, and then fine-tuned the network with back-propagation. \cite{han2015deep} combined pruning, quantization, and Huffman coding to achieve higher compression rate. Gong et al. \cite{gong2014compressing} adopted vector quantization to compress the weighing matrix, which was actually a special case of our approach (apply Q-CNN without error correction to fully-connected layers only).

\section{Experiments}

In this section, we evaluate our quantized CNN framework on two image classification benchmarks, MNIST \cite{lecun1998gradient} and ILSVRC-12 \cite{russakovsky2015imagenet}. For the acceleration of convolutional layers, we compare with:
\vspace{-.5\topsep}
\begin{itemize}
\setlength{\itemsep}{0pt}
\setlength{\parskip}{0pt}
\setlength{\parsep}{0pt}
\item \textbf{CPD} \cite{lebedev2015speeding}: CP-Decomposition;
\item \textbf{GBD} \cite{lebedev2015fast}: Group-wise Brain Damage;
\item \textbf{LANR} \cite{zhang2015accelerating}: Low-rank Approximation of Non-linear Responses.
\end{itemize}
\vspace{-.5\topsep}
and for the compression of fully-connected layers, we compare with the following approaches:
\vspace{-.5\topsep}
\begin{itemize}
\setlength{\itemsep}{0pt}
\setlength{\parskip}{0pt}
\setlength{\parsep}{0pt}
\item \textbf{RER} \cite{ciresan2011high}: Random Edge Removal;
\item \textbf{LRD} \cite{denil2013predicting}: Low-Rank Decomposition;
\item \textbf{DK} \cite{hinton2015distilling}: Dark Knowledge;
\item \textbf{HashNet} \cite{chen2015compressing_icml}: Hashed Neural Nets;
\item \textbf{DPP} \cite{srinivas2015data}: Data-free Parameter Pruning;
\item \textbf{SVD} \cite{denton2014exploiting}: Singular Value Decomposition;
\item \textbf{DFC} \cite{yang2014deep}: Deep Fried Convnets.
\end{itemize}
\vspace{-.5\topsep}
For all above baselines, we use their reported results under the same setting for fair comparison. We report the theoretical speed-up for more consistent results, since the realistic speed-up may be affected by various factors, \eg CPU, cache, and RAM. We compare the theoretical and realistic speed-up in Section \ref{ssc:theoretical_vs_realistic}, and discuss the effect of adopting the BLAS library for acceleration.

Our approaches are denoted as ``Q-CNN'' and ``Q-CNN (EC)'', where the latter one adopts error correction while the former one does not. We implement the optimization process of parameter quantization in MATLAB, and fine-tune the resulting network with Caffe \cite{jia2014caffe}. Additional results of our approach can be found in the supplementary material.

\subsection{Results on MNIST}

The MNIST dataset contains 70k images of hand-written digits, 60k used for training and 10k for testing. To evaluate the compression performance, we pre-train two neural networks, one is 3-layer and another one is 5-layer, where each hidden layer contains 1000 units. Different compression techniques are then adopted to compress these two network, and the results are as depicted in Table \ref{tab:mnist_compression}.

\begin{table}[!ht]
\centering
\caption{Comparison on the compression rates and classification error on MNIST, based on a 3-layer network (784-1000-10) and a 5-layer network (784-1000-1000-1000-10).}
\begin{tabular}{c|c|c|c|c}
\hline
\multirow{2}{*}{Method} & \multicolumn{2}{c|}{3-layer} & \multicolumn{2}{c}{5-layer} \\ \cline{2-5}
 & Compr. & Error & Compr. & Error \\ \hline \hline
Original & - & 1.35\% & - & 1.12\% \\ \hline \hline
RER \cite{ciresan2011high} & 8$\times$ & 2.19\% & 8$\times$ & 1.24\% \\ \hline
LRD \cite{denil2013predicting} & 8$\times$ & 1.89\% & 8$\times$ & 1.77\% \\ \hline
DK \cite{hinton2015distilling} & 8$\times$ & 1.71\% & 8$\times$ & 1.26\% \\ \hline
HashNets \cite{chen2015compressing_icml} & 8$\times$ & 1.43\% & 8$\times$ & 1.22\% \\ \hline \hline
Q-CNN & 12.1$\times$ & 1.42\% & 13.4$\times$ & 1.34\% \\ \hline
Q-CNN (EC) & 12.1$\times$ & \textbf{1.39\%} & 13.4$\times$ & \textbf{1.19\%} \\ \hline
\end{tabular}
\label{tab:mnist_compression}
\end{table}

In our Q-CNN framework, the trade-off between accuracy and efficiency is controlled by $M$ (number of subspaces) and $K$ (number of sub-codewrods in each subspace). Since $M = C_{s} / C'_{s}$ is determined once $C'_{s}$ is given, we tune $( C'_{s}, K )$ to adjust the quantization precision. In Table \ref{tab:mnist_compression}, we set the hyper-parameters as $C'_{s} = 4$ and $K = 32$.

From Table \ref{tab:mnist_compression}, we observe that our Q-CNN (EC) approach offers higher compression rates with less performance degradation than all baselines for both networks. The error correction scheme is effective in reducing the accuracy loss, especially for deeper networks (5-layer). Also, we find the performance of both Q-CNN and Q-CNN (EC) quite stable, as the standard deviation of five random runs is merely 0.05\%. Therefore, we report the single-run performance in the remaining experiments.

\subsection{Results on ILSVRC-12}

The ILSVRC-12 benchmark consists of over one million training images drawn from 1000 categories, and a disjoint validation set of 50k images. We report both the top-1 and top-5 classification error rates on the validation set, using single-view testing (central patch only).

We demonstrate our approach on four convolutional networks: AlexNet \cite{krizhevsky2012imagenet}, CaffeNet \cite{jia2014caffe}, CNN-S \cite{chatfield2014return}, and VGG-16 \cite{simonyan2015very}. The first two models have been adopted in several related works, and therefore are included for comparison. CNN-S and VGG-16 use a either wider or deeper structure for better classification accuracy, and are included here to prove the scalability of our approach. We compare all these networks' computation and storage overhead in Table \ref{tab:alexnet_caffenet_cnns_vgg16_stat}, together with their classification error rates on ILSVRC-12.

\begin{table}[!ht]
\centering
\small
\caption{Comparison on the test-phase computation overhead (FLOPs), storage consumption (Bytes), and classification error rates (Top-1/5 Err.) of AlexNet, CaffeNet, CNN-S, and VGG-16.}
\begin{tabular}{c|c|c|c|c}
\hline
Model    & FLOPs & Bytes & Top-1 Err. & Top-5 Err. \\ \hline \hline
AlexNet  & 7.29e+8  & 2.44e+8 & 42.78\% & 19.74\% \\ \hline
CaffeNet & 7.27e+8  & 2.44e+8 & 42.53\% & 19.59\% \\ \hline
CNN-S    & 2.94e+9  & 4.12e+8 & 37.31\% & 15.82\% \\ \hline
VGG-16   & 1.55e+10 & 5.53e+8 & 28.89\% & 10.05\% \\ \hline
\end{tabular}
\label{tab:alexnet_caffenet_cnns_vgg16_stat}
\end{table}

\subsubsection{Quantizing the Convolutional Layer}

To begin with, we quantize the second convolutional layer of AlexNet, which is the most time-consuming layer during the test-phase. In Table \ref{tab:alexnet_conv_2nd}, we report the performance under several $( C'_{s}, K )$ settings, comparing with two baseline methods, CPD \cite{lebedev2015speeding} and GBD \cite{lebedev2015fast}.

\begin{table}[!ht]
\centering
\scriptsize
\caption{Comparison on the speed-up rates and the increase of top-1/5 error rates for accelerating the second convolutional layer in AlexNet, with or without fine-tuning (FT). The hyper-parameters of Q-CNN, $C'_{s}$ and $K$, are as specified in the ``Para.'' column.}
\begin{tabular}{c|c|c|c|c|c|c}
\hline
\multirow{2}{*}{Method} & \multirow{2}{*}{Para.} & \multirow{2}{*}{Speed-up} & \multicolumn{2}{c|}{Top-1 Err. $\uparrow$} & \multicolumn{2}{c}{Top-5 Err. $\uparrow$} \\ \cline{4-7}
 & & & No FT & FT & No FT & FT \\ \hline \hline
\multirow{3}{*}{CPD} & - & 3.19$\times$ & - & - & 0.94\% & 0.44\% \\ \cline{2-7}
 & - & 4.52$\times$ & - & - & 3.20\% & 1.22\% \\ \cline{2-7}
 & - & 6.51$\times$ & - & - & 69.06\% & 18.63\% \\ \hline
\multirow{3}{*}{GBD} & - & 3.33$\times$ & 12.43\% & 0.11\% & - & - \\ \cline{2-7}
 & - & 5.00$\times$ & 21.93\% & 0.43\% & - & - \\ \cline{2-7}
 & - & 10.00$\times$ & 48.33\% & 1.13\% & - & - \\ \hline \hline
\multirow{4}{*}{Q-CNN} & 4/64 & 3.70$\times$ & 10.55\% & 1.63\% & 8.97\% & 1.37\% \\ \cline{2-7}
 & 6/64 & 5.36$\times$ & 15.93\% & 2.90\% & 14.71\% & 2.27\% \\ \cline{2-7}
 & 6/128 & 4.84$\times$ & 10.62\% & 1.57\% & 9.10\% & 1.28\% \\ \cline{2-7}
 & 8/128 & 6.06$\times$ & 18.84\% & 2.91\% & 18.05\% & 2.66\% \\ \hline
\multirow{4}{*}{\specialcell{Q-CNN \\ (EC)}} & 4/64 & 3.70$\times$ & 0.35\% & 0.20\% & 0.27\% & 0.17\% \\ \cline{2-7}
 & 6/64 & 5.36$\times$ & 0.64\% & 0.39\% & 0.50\% & 0.40\% \\ \cline{2-7}
 & 6/128 & 4.84$\times$ & 0.27\% & 0.11\% & 0.34\% & 0.21\% \\ \cline{2-7}
 & 8/128 & 6.06$\times$ & 0.55\% & 0.33\% & 0.50\% & 0.31\% \\ \hline
\end{tabular}
\label{tab:alexnet_conv_2nd}
\end{table}

From Table \ref{tab:alexnet_conv_2nd}, we discover that with a large speed-up rate (over 4$\times$), the performance loss of both CPD and GBD become severe, especially before fine-tuning. The naive parameter quantization method also suffers from the similar problem. By incorporating the idea of error correction, our Q-CNN model achieves up to 6$\times$ speed-up with merely 0.6\% drop in accuracy, even without fine-tuning. The accuracy loss can be further reduced after fine-tuning the subsequent layers. Hence, it is more effective to minimize the estimation error of each layer's response than minimize the quantization error of network parameters.

\begin{table*}[!ht]
\centering
\caption{Comparison on the speed-up/compression rates and the increase of top-1/5 error rates for accelerating all the convolutional layers in AlexNet and VGG-16.}
\begin{tabular}{c|c|c|c|c|c|c|c|c}
\hline
\multirow{2}{*}{Model} & \multirow{2}{*}{Method} & \multirow{2}{*}{Para.} & \multirow{2}{*}{Speed-up} & \multirow{2}{*}{Compression} & \multicolumn{2}{c|}{Top-1 Err. $\uparrow$} & \multicolumn{2}{c}{Top-5 Err. $\uparrow$} \\ \cline{6-9}
 & & & & & No FT & FT & No FT & FT \\ \hline \hline
\multirow{4}{*}{AlexNet} & \multirow{4}{*}{\specialcell{Q-CNN \\ (EC)}} & 4/64 & 3.32$\times$ & 10.58$\times$ & 1.33\% & - & 0.94\% & - \\ \cline{3-9}
 & & 6/64 & 4.32$\times$ & 14.32$\times$ & 2.32\% & - & 1.90\% & - \\ \cline{3-9}
 & & 6/128 & 3.71$\times$ & 10.27$\times$ & 1.44\% & 0.13\% & 1.16\% & 0.36\% \\ \cline{3-9}
 & & 8/128 & 4.27$\times$ & 12.08$\times$ & 2.25\% & 0.99\% & 1.64\% & 0.60\% \\ \hline
\multirow{2}{*}{VGG-16} & LANR \cite{zhang2015accelerating} & - & 4.00$\times$ & 2.73$\times$ & - & - & 0.95\% & 0.35\% \\ \cline{2-9}
 & Q-CNN (EC) & 6/128 & 4.06$\times$ & 14.40$\times$ & 3.04\% & 1.06\% & 1.83\% & 0.45\% \\ \hline
\end{tabular}
\label{tab:alexnet_vgg16_conv_all}
\end{table*}

Next, we take one step further and attempt to speed-up all the convolutional layers in AlexNet with Q-CNN (EC). We fix the quantization hyper-parameters $( C'_{s}, K )$ across all layers. From Table \ref{tab:alexnet_vgg16_conv_all}, we observe that the loss in accuracy grows mildly than the single-layer case. The speed-up rates reported here are consistently smaller than those in Table \ref{tab:alexnet_conv_2nd}, since the acceleration effect is less significant for some layers (i.e. ``conv\_4'' and ``conv\_5''). For AlexNet, our Q-CNN model ($C'_{s} = 8, K = 128$) can accelerate the computation of all the convolutional layers by a factor of 4.27$\times$, while the increase in the top-1 and top-5 error rates are no more than 2.5\%. After fine-tuning the remaining fully-connected layers, the performance loss can be further reduced to less than 1\%.

In Table \ref{tab:alexnet_vgg16_conv_all}, we also report the comparison against LANR \cite{zhang2015accelerating} on VGG-16. For the similar speed-up rate (4$\times$), their approach outperforms ours in the top-5 classification error (an increase of 0.95\% against 1.83\%). After fine-tuning, the performance gap is narrowed down to 0.35\% against 0.45\%. At the same time, our approach offers over $14\times$ compression of parameters in convolutional layers, much larger than theirs $2.7\times$ compression\footnotemark. Therefore, our approach is effective in accelerating and compressing networks with many convolutional layers, with only minor performance loss.

\footnotetext{The compression effect of their approach was not explicitly discussed in the paper; we estimate the compression rate based on their description.}

\subsubsection{Quantizing the Fully-connected Layer}

For demonstration, we first compress parameters in a single fully-connected layer. In CaffeNet, the first fully-connected layer possesses over 37 million parameters ($9216 \times 4096$), more than 60\% of whole network parameters. Our Q-CNN approach is adopted to quantize this layer and the results are as reported in Table \ref{tab:caffenet_fcnt_1st}. The performance loss of our Q-CNN model is negligible (within 0.4\%), which is much smaller than baseline methods (DPP and SVD). Furthermore, error correction is effective in preserving the classification accuracy, especially under a higher compression rate.

\begin{table}[!ht]
\centering
\scriptsize
\caption{Comparison on the compression rates and the increase of top-1/5 error rates for compressing the first fully-connected layer in CaffeNet, without fine-tuning.}
\begin{tabular}{c|c|c|c|c}
\hline
Method & Para. & Compression & Top-1 Err. $\uparrow$ & Top-5 Err. $\uparrow$ \\ \hline \hline
\multirow{4}{*}{DPP} & - & 1.19$\times$ & 0.16\% & - \\ \cline{2-5}
 & - & 1.47$\times$ & 1.76\% & - \\ \cline{2-5}
 & - & 1.91$\times$ & 4.08\% & - \\ \cline{2-5}
 & - & 2.75$\times$ & 9.68\% & - \\ \hline
\multirow{4}{*}{SVD} & - & 1.38$\times$ & 0.03\% & -0.03\% \\ \cline{2-5}
 & - & 2.77$\times$ & 0.07\% & 0.07\% \\ \cline{2-5}
 & - & 5.54$\times$ & 0.36\% & 0.19\% \\ \cline{2-5}
 & - & 11.08$\times$ & 1.23\% & 0.86\% \\ \hline \hline
\multirow{4}{*}{Q-CNN} & 2/16 & 15.06$\times$ & 0.19\% & 0.19\% \\ \cline{2-5}
 & 3/16 & 21.94$\times$ & 0.35\% & 0.28\% \\ \cline{2-5}
 & 3/32 & 16.70$\times$ & 0.18\% & 0.12\% \\ \cline{2-5}
 & 4/32 & 21.33$\times$ & 0.28\% & 0.16\% \\ \hline
\multirow{4}{*}{\specialcell{Q-CNN \\ (EC)}} & 2/16 & 15.06$\times$ & 0.10\% & 0.07\% \\ \cline{2-5}
 & 3/16 & 21.94$\times$ & 0.18\% & 0.03\% \\ \cline{2-5}
 & 3/32 & 16.70$\times$ & 0.14\% & 0.11\% \\ \cline{2-5}
 & 4/32 & 21.33$\times$ & 0.16\% & 0.12\% \\ \hline
\end{tabular}
\label{tab:caffenet_fcnt_1st}
\end{table}

Now we evaluate our approach's performance for compressing all the fully-connected layers in CaffeNet in Table \ref{tab:caffenet_fcnt_all}. The third layer is actually the combination of 1000 classifiers, and is more critical to the classification accuracy. Hence, we adopt a much more fine-grained hyper-parameter setting ($C'_{s} = 1, K = 16$) for this layer. Although the speed-up effect no longer exists, we can still achieve around 8$\times$ compression for the last layer.

\begin{table}[!ht]
\centering
\scriptsize
\caption{Comparison on the compression rates and the increase of top-1/5 error rates for compressing all the fully-connected layers in CaffeNet. Both SVD and DFC are fine-tuned, while Q-CNN and Q-CNN (EC) are not fine-tuned.}
\begin{tabular}{c|c|c|c|c}
\hline
Method & Para. & Compression & Top-1 Err. $\uparrow$ & Top-5 Err. $\uparrow$ \\ \hline \hline
\multirow{2}{*}{SVD} & - & 1.26$\times$ & 0.14\% & - \\ \cline{2-5}
 & - & 2.52$\times$ & 1.22\% & - \\ \hline
\multirow{2}{*}{DFC} & - & 1.79$\times$ & -0.66\% & - \\ \cline{2-5}
 & - & 3.58$\times$ & 0.31\% & - \\ \hline \hline
\multirow{4}{*}{Q-CNN} & 2/16 & 13.96$\times$ & 0.28\% & 0.29\% \\ \cline{2-5}
 & 3/16 & 19.14$\times$ & 0.70\% & 0.47\% \\ \cline{2-5}
 & 3/32 & 15.25$\times$ & 0.44\% & 0.34\% \\ \cline{2-5}
 & 4/32 & 18.71$\times$ & 0.75\% & 0.59\% \\ \hline
\multirow{4}{*}{\specialcell{Q-CNN \\ (EC)}} & 2/16 & 13.96$\times$ & 0.31\% & 0.30\% \\ \cline{2-5}
 & 3/16 & 19.14$\times$ & 0.59\% & 0.47\% \\ \cline{2-5}
 & 3/32 & 15.25$\times$ & 0.31\% & 0.27\% \\ \cline{2-5}
 & 4/32 & 18.71$\times$ & 0.57\% & 0.39\% \\ \hline
\end{tabular}
\label{tab:caffenet_fcnt_all}
\end{table}

From Table \ref{tab:caffenet_fcnt_all}, we discover that with less than 1\% drop in accuracy, Q-CNN achieves high compression rates ($12 \sim 20\times$), much larger than that of SVD\footnotemark and DFC ($<4\times$). Again, Q-CNN with error correction consistently outperforms the naive Q-CNN approach as adopted in \cite{gong2014compressing}.

\footnotetext{In Table \ref{tab:caffenet_fcnt_1st}, SVD means replacing the weighting matrix with the multiplication of two low-rank matrices; in Table \ref{tab:caffenet_fcnt_all}, SVD means fine-tuning the network after the low-rank matrix decomposition.}

\subsubsection{Quantizing the Whole Network}

So far, we have evaluated the performance of CNN models with either convolutional or fully-connected layers quantized. Now we demonstrate the quantization of the whole network with a three-stage strategy. Firstly, we quantize all the convolutional layers with error correction, while fully-connected layers remain untouched. Secondly, we fine-tune fully-connected layers in the quantized network with the ILSVRC-12 training set to restore the classification accuracy. Finally, fully-connected layers in the fine-tuned network are quantized with error correction. We report the performance of our Q-CNN models in Table \ref{tab:alexnet_caffenet_cnns_vgg16_whole_net}.

\begin{table}[!ht]
\centering
\scriptsize
\caption{The speed-up/compression rates and the increase of top-1/5 error rates for the whole CNN model. Particularly, for the quantization of the third fully-connected layer in each network, we let $C'_{s} = 1$ and $K = 16$.}
\begin{tabular}{c|c|c|c|c|c}
\hline
\multirow{2}{*}{Model} & \multicolumn{2}{c|}{Para.} & \multirow{2}{*}{Speed-up} & \multirow{2}{*}{Compression} & \multirow{2}{*}{Top-1/5 Err. $\uparrow$} \\ \cline{2-3}
& Conv. & FCnt. & & & \\ \hline \hline
\multirow{2}{*}{AlexNet} & 8/128 & 3/32 & 4.05$\times$ & 15.40$\times$ & 1.38\% / 0.84\% \\ \cline{2-6}
 & 8/128 & 4/32 & 4.15$\times$ & 18.76$\times$ & 1.46\% / 0.97\% \\ \hline
 \multirow{2}{*}{CaffeNet} & 8/128 & 3/32 & 4.04$\times$ & 15.40$\times$ & 1.43\% / 0.99\% \\ \cline{2-6}
 & 8/128 & 4/32 & 4.14$\times$ & 18.76$\times$ & 1.54\% / 1.12\% \\ \hline
 \multirow{2}{*}{CNN-S} & 8/128 & 3/32 & 5.69$\times$ & 16.32$\times$ & 1.48\% / 0.81\% \\ \cline{2-6}
 & 8/128 & 4/32 & 5.78$\times$ & 20.16$\times$ & 1.64\% / 0.85\% \\ \hline
 \multirow{2}{*}{VGG-16} & 6/128 & 3/32 & 4.05$\times$ & 16.55$\times$ & 1.22\% / 0.53\% \\ \cline{2-6}
 & 6/128 & 4/32 & 4.06$\times$ & 20.34$\times$ & 1.35\% / 0.58\% \\ \hline
\end{tabular}
\label{tab:alexnet_caffenet_cnns_vgg16_whole_net}
\end{table}

For convolutional layers, we let $C'_{s} = 8$ and $K = 128$ for AlexNet, CaffeNet, and CNN-S, and let $C'_{s} = 6$ and $K = 128$ for VGG-16, to ensure roughly $4 \sim 6 \times$ speed-up for each network. Then we vary the hyper-parameter settings in fully-connected layers for different compression levels. For the former two networks, we achieve 18$\times$ compression with about 1\% loss in the top-5 classification accuracy. For CNN-S, we achieve 5.78$\times$ speed-up and 20.16$\times$ compression, while the top-5 classification accuracy drop is merely 0.85\%. The result on VGG-16 is even more encouraging: with 4.06$\times$ speed-up and 20.34$\times$, the increase of top-5 error rate is only 0.58\%. Hence, our proposed Q-CNN framework can improve the efficiency of convolutional networks with minor performance loss, which is acceptable in many applications.

\subsection{Results on Mobile Devices}

We have developed an Android application to fulfill CNN-based image classification on mobile devices, based on our Q-CNN framework. The experiments are carried out on a Huawei\textsuperscript{\textregistered} Mate 7 smartphone, equipped with an 1.8GHz Kirin 925 CPU. The test-phase computation is carried out on a single CPU core, without GPU acceleration.

\begin{table}[!ht]
\centering
\scriptsize
\caption{Comparison on the time, storage, memory consumption, and top-5 classification error rates of the original and quantized AlexNet and CNN-S.}
\begin{tabular}{c|c|c|c|c|c}
\hline
\multicolumn{2}{c|}{Model} & Time & Storage & Memory & Top-5 Err. \\ \hline \hline
\multirow{2}{*}{AlexNet} & CNN & ~~2.93s & 232.56MB & 264.74MB & 19.74\%\\ \cline{2-6}
 & \textbf{Q-CNN} & \textbf{~~0.95s} & \textbf{~~12.60MB} & \textbf{~~74.65MB} & \textbf{20.70\%} \\ \hline
\multirow{2}{*}{CNN-S} & CNN & 10.58s & 392.57MB & 468.90MB & 15.82\% \\ \cline{2-6}
 & \textbf{Q-CNN} & \textbf{~~2.61s} & \textbf{~~20.13MB} & \textbf{129.49MB} & \textbf{16.68\%} \\ \hline
\end{tabular}
\label{tab:mobile_devices}
\end{table}

In Table \ref{tab:mobile_devices}, we compare the computation efficiency and classification accuracy of the original and quantized CNN models. Our Q-CNN framework achieves 3$\times$ speed-up for AlexNet, and 4$\times$ speed-up for CNN-S. What's more, we compress the storage consumption by 20 $\times$, and the required run-time memory is only one quarter of the original model. At the same time, the loss in the top-5 classification accuracy is no more than 1\%. Therefore, our proposed approach improves the run-time efficiency in multiple aspects, making the deployment of CNN models become tractable on mobile platforms.

\subsection{Theoretical vs. Realistic Speed-up}
\label{ssc:theoretical_vs_realistic}

In Table \ref{tab:theoretical_vs_realistic}, we compare the theoretical and realistic speed-up on AlexNet. The BLAS \cite{whaley2004minimizing} library is used in Caffe \cite{jia2014caffe} to accelerate the matrix multiplication in convolutional and fully-connected layers. However, it may not always be an option for mobile devices. Therefore, we measure the run-time speed under two settings, \ie with BLAS enabled or disabled. The realistic speed-up is slightly lower with BLAS on, indicating that Q-CNN does not benefit as much from BLAS as that of CNN. Other optimization techniques, \eg SIMD, SSE, and AVX \cite{intel2016intel}, may further improve our realistic speed-up, and shall be explored in the future.

\begin{table}[!ht]
\centering
\scriptsize
\caption{Comparison on the theoretical and realistic speed-up on AlexNet (CPU only, single-threaded). Here we use the ATLAS library, which is the default BLAS choice in Caffe \cite{jia2014caffe}.}
\begin{tabular}{c|c|c|c|c|c|c}
\hline
\multirow{2}{*}{BLAS} & \multicolumn{2}{c|}{FLOPs} & \multicolumn{2}{c|}{Time (ms)} & \multicolumn{2}{c}{Speed-up} \\ \cline{2-7}
 & CNN & Q-CNN & CNN & Q-CNN & Theo. & Real. \\ \hline \hline
Off & \multirow{2}{*}{7.29e+8} & \multirow{2}{*}{1.75e+8} & 321.10 & 75.62 & \multirow{2}{*}{4.15$\times$} & 4.25$\times$ \\ \cline{1-1} \cline{4-5} \cline{7-7}
On & & & 167.79\footnotemark & 55.35 & & 3.03$\times$ \\ \hline
\end{tabular}
\label{tab:theoretical_vs_realistic}
\end{table}
\footnotetext{This is Caffe's run-time speed. The code for the other three settings is on \url{https://github.com/jiaxiang-wu/quantized-cnn}.}

\section{Conclusion}

In this paper, we propose a unified framework to simultaneously accelerate and compress convolutional neural networks. We quantize network parameters to enable efficient test-phase computation. Extensive experiments are conducted on MNIST and ILSVRC-12, and our approach achieves outstanding speed-up and compression rates, with only negligible loss in the classification accuracy.

\section{Acknowledgement}
This work was supported in part by National Natural Science Foundation of China (Grant No. 61332016), and 863 program (Grant No. 2014AA015105).

{
\footnotesize
\bibliographystyle{ieee}
\bibliography{bibtex_database}
}

\newpage

\section*{Appendix A: Additional Results}

In the submission, we report the performance after quantizing all the convolutional layers in AlexNet, and quantizing all the full-connected layers in CaffeNet. Here, we present experimental results for some other settings.

\subsection*{Quantizing Convolutional Layers in CaffeNet}

We quantize all the convolutional layers in CaffeNet, and the results are as demonstrated in Table \ref{tab:caffenet_conv_all}. Furthermore, we fine-tune the quantized CNN model learned with error correction ($C'_{s} = 8, K = 128$), and the increase of top-1/5 error rates are 1.15\% and 0.75\%, compared to the original CaffeNet.

\begin{table}[!ht]
\centering
\footnotesize
\caption{Comparison on the speed-up rates and the increase of top-1/5 error rates for accelerating all the convolutional layers in CaffeNet, without fine-tuning.}
\begin{tabular}{c|c|c|c|c}
\hline
Method & Para. & Speed-up & Top-1 Err. $\uparrow$ & Top-5 Err. $\uparrow$ \\ \hline \hline
\multirow{4}{*}{Q-CNN} & 4/64 & 3.32$\times$ & 18.69\% & 16.73\% \\ \cline{2-5}
 & 6/64 & 4.32$\times$ & 32.84\% & 33.55\% \\ \cline{2-5}
 & 6/128 & 3.71$\times$ & 20.08\% & 18.31\% \\ \cline{2-5}
 & 8/128 & 4.27$\times$ & 35.48\% & 37.82\% \\ \hline
\multirow{4}{*}{\specialcell{Q-CNN \\ (EC)}} & 4/64 & 3.32$\times$ & 1.22\% & 0.97\% \\ \cline{2-5}
 & 6/64 & 4.32$\times$ & 2.44\% & 1.83\% \\ \cline{2-5}
 & 6/128 & 3.71$\times$ & 1.57\% & 1.12\% \\ \cline{2-5}
 & 8/128 & 4.27$\times$ & 2.30\% & 1.71\% \\ \hline
\end{tabular}
\label{tab:caffenet_conv_all}
\end{table}

\subsection*{Quantizing Convolutional Layers in CNN-S}

We quantize all the convolutional layers in CNN-S, and the results are as demonstrated in Table \ref{tab:vggcnns_conv_all}. Furthermore, we fine-tune the quantized CNN model learned with error correction ($C'_{s} = 8, K = 128$), and the increase of top-1/5 error rates are 1.24\% and 0.63\%, compared to the original CNN-S.

\begin{table}[!ht]
\centering
\footnotesize
\caption{Comparison on the speed-up rates and the increase of top-1/5 error rates for accelerating all the convolutional layers in CNN-S, without fine-tuning.}
\begin{tabular}{c|c|c|c|c}
\hline
Method & Para. & Speed-up & Top-1 Err. $\uparrow$ & Top-5 Err. $\uparrow$ \\ \hline \hline
\multirow{4}{*}{Q-CNN} & 4/64 & 3.69$\times$ & 19.87\% & 16.77\% \\ \cline{2-5}
 & 6/64 & 5.17$\times$ & 45.74\% & 48.67\% \\ \cline{2-5}
 & 6/128 & 4.78$\times$ & 27.86\% & 25.09\% \\ \cline{2-5}
 & 8/128 & 5.92$\times$ & 46.18\% & 50.26\% \\ \hline
\multirow{4}{*}{\specialcell{Q-CNN \\ (EC)}} & 4/64 & 3.69$\times$ & 1.60\% & 0.92\% \\ \cline{2-5}
 & 6/64 & 5.17$\times$ & 3.49\% & 2.32\% \\ \cline{2-5}
 & 6/128 & 4.78$\times$ & 2.07\% & 1.32\% \\ \cline{2-5}
 & 8/128 & 5.92$\times$ & 3.42\% & 2.17\% \\ \hline
\end{tabular}
\label{tab:vggcnns_conv_all}
\end{table}

\subsection*{Quantizing Fully-connected Layers in AlexNet}

We quantize all the fully-connected layers in AlexNet, and the results are as demonstrated in Table \ref{tab:alexnet_fcnt_all}.

\begin{table}[!ht]
\centering
\footnotesize
\caption{Comparison on the compression rates and the increase of top-1/5 error rates for compressing all the fully-connected layers in AlexNet, without fine-tuning.}
\begin{tabular}{c|c|c|c|c}
\hline
Method & Para. & Compression & Top-1 Err. $\uparrow$ & Top-5 Err. $\uparrow$ \\ \hline \hline
\multirow{4}{*}{Q-CNN} & 2/16 & 13.96$\times$ & 0.25\% & 0.27\% \\ \cline{2-5}
 & 3/16 & 19.14$\times$ & 0.77\% & 0.64\% \\ \cline{2-5}
 & 3/32 & 15.25$\times$ & 0.54\% & 0.33\% \\ \cline{2-5}
 & 4/32 & 18.71$\times$ & 0.71\% & 0.69\% \\ \hline
\multirow{4}{*}{\specialcell{Q-CNN \\ (EC)}} & 2/16 & 13.96$\times$ & 0.14\% & 0.20\% \\ \cline{2-5}
 & 3/16 & 19.14$\times$ & 0.40\% & 0.22\% \\ \cline{2-5}
 & 3/32 & 15.25$\times$ & 0.40\% & 0.21\% \\ \cline{2-5}
 & 4/32 & 18.71$\times$ & 0.46\% & 0.38\% \\ \hline
\end{tabular}
\label{tab:alexnet_fcnt_all}
\end{table}

\subsection*{Quantizing Fully-connected Layers in CNN-S}

We quantize all the fully-connected layers in CNN-S, and the results are as demonstrated in Table \ref{tab:vggcnns_fcnt_all}.

\begin{table}[!ht]
\centering
\footnotesize
\caption{Comparison on the compression rates and the increase of top-1/5 error rates for compressing all the fully-connected layers in CNN-S, without fine-tuning.}
\begin{tabular}{c|c|c|c|c}
\hline
Method & Para. & Compression & Top-1 Err. $\uparrow$ & Top-5 Err. $\uparrow$ \\ \hline \hline
\multirow{4}{*}{Q-CNN} & 2/16 & 14.37$\times$ & 0.22\% & 0.07\% \\ \cline{2-5}
 & 3/16 & 20.15$\times$ & 0.45\% & 0.22\% \\ \cline{2-5}
 & 3/32 & 15.79$\times$ & 0.21\% & 0.11\% \\ \cline{2-5}
 & 4/32 & 19.66$\times$ & 0.35\% & 0.27\% \\ \hline
\multirow{4}{*}{\specialcell{Q-CNN \\ (EC)}} & 2/16 & 14.37$\times$ & 0.36\% & 0.14\% \\ \cline{2-5}
 & 3/16 & 20.15$\times$ & 0.43\% & 0.24\% \\ \cline{2-5}
 & 3/32 & 15.79$\times$ & 0.29\% & 0.11\% \\ \cline{2-5}
 & 4/32 & 19.66$\times$ & 0.56\% & 0.27\% \\ \hline
\end{tabular}
\label{tab:vggcnns_fcnt_all}
\end{table}

\section*{Appendix B: Optimization in Section \ref{sss:err_corr_conv}}

Assume we have $N$ images to learn the quantization of a convolutional layer. For image $I_{n}$, we denote its input feature maps as $S_{n} \in \mathbb{R}^{d_{s} \times d_{s} \times C_{s}}$ and response feature maps as $T_{n} \in \mathbb{R}^{d_{t} \times d_{t} \times C_{t}}$, where $d_{s}, d_{t}$ are the spatial sizes and $C_{s}, C_{t}$ are the number of feature map channels. We use $p_{s}$ and $p_{t}$ to denote the spatial location in the input and response feature maps. The spatial location in the convolutional kernels is denoted as $p_{k}$.

To learn quantization with error correction for the convolutional layer, we attempt to optimize:
\small
\begin{equation}
\min\limits_{\{ D^{( m )} \}, \{ B_{p_{k}}^{( m )} \}} \sum_{n, p_{t}} \left\| \sum_{( p_{k}, p_{s} )} \sum_{m} ( D^{( m )} B_{p_{k}}^{( m )} )^{T} S_{n, p_{s}}^{( m )} - T_{n, p_{t}} \right\|_{F}^{2}
\end{equation}
\normalsize
where $D^{m}$ is the $m$-th sub-codebook, and $B_{p_{k}}^{( m )}$ is the corresponding sub-codeword assignment indicator for the convolutional kernels at spatial location $p_{k}$.

Similar to the fully-connected layer, we adopt a block coordinate descent approach to solve this optimization problem. For the $m$-th subspace, we firstly define its residual feature map as:
\begin{equation}
R_{n, p_{t}}^{( m )} = T_{n, p_{t}} - \sum_{( p_{k}, p_{s} )} \sum_{m' \neq m} ( D^{( m' )} B_{p_{k}}^{( m' )} )^{T} S_{n, p_{s}}^{( m' )}
\end{equation}
and then the optimization in the $m$-th subspace can be re-formulated as:
\begin{equation}
\min\limits_{D^{( m )}, \{ B_{p_{k}}^{( m )} \}} \sum_{n, p_{t}} \left\| \sum_{( p_{k}, p_{s} )} ( D^{( m )} B_{p_{k}}^{( m )} )^{T} S_{n, p_{s}}^{( m )} - R_{n, p_{t}}^{( m )} \right\|_{F}^{2}
\end{equation}

\textbf{Update \boldmath$D^{( m )}$.} With the assignment indicator $\{ B_{p_{k}}^{( m )} \}$ fixed, we let:
\begin{equation}
L_{k, p_{k}} = \{ c_{t} | B_{p_{k}}^{( m )} ( k, c_{t} ) = 1 \}
\end{equation}

We greedily update each sub-codeword in the $m$-th sub-codebook $D^{( m )}$ in a sequential style. For the $k$-th sub-codeword, we compute the corresponding residual feature map as:
\small
\begin{equation}
Q_{n, p_{t}, k}^{( m )} ( c_{t} ) = R_{n, p_{t}}^{( m )} ( c_{t} ) - \sum_{( p_{k}, p_{s} )} \sum_{k' \neq k} \sum_{c_{t} \in L_{k', p_{k}}} D_{k'}^{( m )^{T}} S_{n, p_{s}}^{( m )}
\end{equation}
\normalsize
and then we can alternatively optimize:
\begin{equation}
\min\limits_{D_{k}^{( m )}} \sum_{n, p_{t}} \left\| \sum_{( p_{k}, p_{s} )} \sum_{c_{t} \in L_{k, p_{k}}}D_{k}^{( m )^{T}} S_{n, p_{s}}^{( m )} - Q_{n, p_{t}, k}^{( m )} ( c_{t} ) \right\|_{F}^{2}
\end{equation}
which can be transformed into a least square problem. By solving it, we can update the $k$-th sub-codeword.

\textbf{Update \boldmath$\{ B_{p_{k}}^{( m )} \}$.} We greedily update the sub-codeword assignment at each spatial location in the convolutional kernels in a sequential style. For the spatial location $p_{k}$, we compute the corresponding residual feature map as:
\begin{equation}
P_{n, p_{t}, p_{k}}^{( m )} = R_{n, p_{t}}^{( m )} - \sum_{\substack{( p'_{k}, p'_{s} ) \\ p_{k} \neq p_{k}}} ( D^{( m )} B_{p'_{k}}^{( m )} )^{T} S_{n, p'_{s}}^{( m )}
\end{equation}
and then the optimization can be re-written as:
\begin{equation}
\min\limits_{B_{p_{k}}^{( m )}} \sum_{n, p_{t}} \left\| ( D^{( m )} B_{p_{k}}^{( m )} )^{T} S_{n, p_{s}}^{( m )} - P_{n, p_{t}, p_{k}}^{( m )} \right\|_{F}^{2}
\end{equation}
Since $B_{p_{k}}^{( m )} \in \left\{ 0, 1 \right\}^{K}$ is an indicator vector (only one non-zero entry), we can exhaustively try all sub-codewords and select the optimal one that minimize the objective function.

\end{document}